\documentclass[letterpaper, 10 pt, conference]{ieeeconf}  

\IEEEoverridecommandlockouts                              

\usepackage{amsmath,amsfonts}
\usepackage{algpseudocode}
\usepackage[vlined,ruled]{algorithm2e}
\usepackage{array}
\usepackage{textcomp}
\usepackage{stfloats}
\usepackage{url}
\usepackage{verbatim}
\usepackage{graphicx}
\usepackage[outdir=./]{epstopdf}
\usepackage{cite}
\usepackage{mathtools}
\usepackage{float}
\usepackage{graphicx}
\usepackage{amsmath}
\usepackage{arydshln}
\usepackage{tensor}
\usepackage{multirow}
\usepackage{booktabs}
\usepackage{gensymb}
\usepackage{amssymb}
\usepackage{bbding}
\usepackage{bm}
\usepackage{subcaption}
\usepackage{siunitx}
\usepackage[colorlinks,linkcolor=black]{hyperref}
\usepackage[switch]{lineno}
\usepackage{relsize}
\usepackage{makecell}

\usepackage{xcolor}  
\definecolor{bevtopo_color}{RGB}{255,0,0}
\definecolor{NetVLAD_color}{RGB}{164,217,187}
\definecolor{LoST_color}{RGB}{239,129,131}
\definecolor{R2Former_color}{RGB}{105,158,212}

\newcommand{\eref}[1]{Eq.~(\ref{#1})}
\newcommand{\fref}[1]{Fig.~\ref{#1}}
\newcommand{\sref}[1]{Sec.~\ref{#1}}
\newcommand{\tref}[1]{Table~\ref{#1}}
\newcommand{\aref}[1]{Alg.~\ref{#1}}

\title{\LARGE \bf
A Skeleton-Based Topological Planner for Exploration 
\\ in Complex Unknown Environments
}

\author{Haochen Niu, Xingwu Ji, Lantao Zhang, Fei Wen, Rendong Ying and Peilin Liu* %
\thanks{This work was supported by the National Natural Science Foundation of China (No. 62276166) and the STI 2030-Major Projects (No. 2022ZD0208700).
All authors are with the Brain-Inspired Application Technology Center (BATC), School of Electronic Information and Electrical Engineering, Shanghai Jiao Tong University, Shanghai 200240, China (email: \{haochen\_niu, jixingwu, swagger, wenfei, rdying, liupeilin\}@sjtu.edu.cn).}%
}

\begin{document}

\maketitle
\thispagestyle{empty}
\pagestyle{empty}

\begin{abstract}

The capability of autonomous exploration in complex, unknown environments is important in many robotic applications. While recent research on autonomous exploration have achieved much progress, there are still limitations, e.g., existing methods relying on greedy heuristics or optimal path planning are often hindered by repetitive paths and high computational demands.
To address such limitations, we propose a novel exploration framework that utilizes the global topology information of observed environment to improve exploration efficiency while reducing computational overhead.
Specifically, global information is utilized based on a skeletal topological graph representation of the environment geometry.
We first propose an incremental skeleton extraction method based on wavefront propagation, based on which we then design an approach to generate a lightweight topological graph that can effectively capture the environment's structural characteristics.
Building upon this, we introduce a finite state machine that leverages the topological structure to efficiently plan coverage paths, which can substantially mitigate the back-and-forth maneuvers (BFMs) problem.
Experimental results demonstrate the superiority of our method in comparison with state-of-the-art methods.
The source code will be made publicly available at: \url{https://github.com/Haochen-Niu/STGPlanner}.

\end{abstract}

\section{INTRODUCTION}\label{sec:intro}
Recently, autonomous exploration has attracted much research attention due to its importance in many robotic applications such as search and rescue, environmental monitoring, and planetary exploration\cite{lindqvist2024TreeBased}. The primary objective of autonomous exploration is to enable robots to efficiently navigate and map unknown environments without human assistance or prior knowledge, thereby facilitating and enhancing subsequent mission tasks.

A key challenge in autonomous exploration is making optimal decision in environments with incomplete information. Robots must continuously optimize their paths to maximize information acquisition while minimizing resource consumption, such as time, energy, and computational power. 
Early works, such as frontier-based \cite{yamauchi1997Frontierbased, cieslewski2017Rapid, deng2020Robotic} or sampling-based methods \cite{bircher2016Receding, dang2020Graphbased,zhu2021DSVP}, typically rely on greedy heuristics, which direct robots toward areas with the highest immediate information gain. However, due to the lack of a global perspective, these methods may result in low-quality trajectories, particularly in complex environments\cite{tang2023Bubble, kim2023Topological, zhao2024Autonomous}.
For instance, as the frontier of a local region shrinks during exploration, the robot may greedily shift toward another region with higher utility. As the utility in the new region diminishes, the robot returns to the previous region, leading to inefficient back-and-forth maneuvers (BFMs). 

To address this issue, recent methods have integrated global and local information following a coarse-to-fine hierarchical paradigm, which plan globally optimal paths for exploration via constructing formulations akin to the Traveling Salesman Problem (TSP) \cite{cao2021TARE,zhao2023TDLE,luo2024StarSearcher}. 
However, due to the unpredictability of unknown spaces, these methods, based on partially observable information, still retain a greedy nature. Consequently, the BFMs problem can only be partially mitigated. Additionally, as the observed information changes during exploration, 
the optimal path needs to be recurrently replanned,
which incurs a significant computational burden due to solving the TSP formulation frequently.

To address the aforementioned limitations, this work aims to leverage the environment's topology to guide exploration, rather than relying solely on maximum information gain or ``global optimality".
To this end, we propose STGPlanner, a novel autonomous exploration framework based on the skeletal topological graph (STG) of the environment.
Specifically, we develop an incremental skeleton extraction method that captures the environment's topology from its observed geometry, forming the basis for generating the STG.
This process implicitly integrates key elements such as frontiers, viewpoints, information gain, and path planning.
Using this representation, we introduce high-level intention that prioritizes the complete exploration of the current STG branch to substantially mitigate inefficient BFMs.
Building on this concept, a STG-based exploration strategy encapsulated within a finite state machine (FSM) is designed, which enables efficient exploration at a low computational cost.

The main contributions are as follows.
\begin{itemize}
    \item 
    A novel autonomous exploration framework, which utilizes the global topology information of observed environment to effectively address the BFMs problem, based on a STG representation of the environment geometry.

    \item 
    A STG generation method based on incremental skeleton extraction, which can efficiently construct a concise topology of the observed environment with essential exploration information.

    \item 
    A STG-based exploration strategy encapsulated within a FSM, which fully leverages the information of STG to enable efficient exploration decision at a low computational cost.
    
    \item 
    Extensive experimental evaluation showcases the effectiveness of our method in various environments.
\end{itemize} 

\section{RELATED WORK}\label{sec:RW}
In the past decade, autonomous exploration has been extensively studied
and many methods have been proposed. Typically, autonomous exploration involves three key stages: identification of candidates target/actions, utility evaluation, planning and execution \cite{placed2023Survey}.

The work \cite{yamauchi1997Frontierbased} pioneered the frontier-based method, which introduces the concept of frontiers as the boundaries between known and unknown areas and uses frontiers as candidate targets, e.g., select the nearest frontier for exploration.
This seminal work has inspired numerous subsequent improvements.
For example, the work \cite{cieslewski2017Rapid} incorporates field of view consideration to minimize speed variation and maintain high-speed movement.
The work \cite{zhou2023RACER} proposes the frontier information structure (FIS) for rapid frontier detection, while \cite{deng2020Robotic} revisits frontier information gain through an automatic-differentiable function with respect to the robot’s path.

Another class of methods for candidate target selection is sampling-based. 
RH-NBVP \cite{bircher2016Receding} first applies the next best view in a receding horizon manner to autonomous exploration. It spans a rapidly-exploring random tree (RRT) to calculate the information gain of each node via ray-casting and chooses the path with the highest utility, executing only the first step.
Expanding on this foundation, the works \cite{zhu2021DSVP, xu2023Heuristicbased} introduce biased sampling to steer RRT growth towards unexplored regions, thereby improving exploration efficiency. 
Furthermore, \cite{dang2020Graphbased} employs a dual-layer planning architecture and uses a dense rapidly-random graph to identify collision-free paths with maximal information gain.

However, these methods' reliance on greedy decisions often results in low-quality trajectories due to a lack of global perspective \cite{zhao2024Autonomous, chen2023STExplorer}. To counter this, recent approaches integrate global and local information, using TSP formulations to optimize global trajectories based on current information.
For example, the work \cite{huang2023FAEL} merges frontier and sampling methods, counting visible frontiers from sampled viewpoints as the information gain, reducing the computational overhead of ray-casting. 
It then uses the 2-opt algorithm to solve the TSP and find the best viewpoint order. 
Similarly, \cite{tang2023Bubble} uses occlusion-free spheres to improve viewpoint selection, treating it as an asymmetric TSP.
To reduce computational complexity, some works combine a coarse global planning with a fine local planning.
For example, the works \cite{cao2021TARE,zhao2023TDLE,hui2024DPPM} initially segment the map into grids to determine a coarse path via TSP, covering unexplored areas. Then, they  sample viewpoints locally, compute information gain, and reapply TSP to refine the trajectory.
Moreover, the work \cite{luo2024StarSearcher} replaces uniform grid with viewpoints clustering to achieve a more flexible global representation.
Despite the efficiency improvement by incorporating global optimization, BFMs remain unavoidable due to the heuristic utility definitions and the partially observable information during exploration.
Moreover, methods like \cite{tao2023SEER} utilize semantic markers, such as door detection, to prioritize exploration of small regions, which introduce additional computational overhead and are restricted to specific environments.

To address these limitations, we utilize the environment's topology information to guide the robot in prioritizing comprehensive exploration within the current region, thereby reducing BFMs.
Central to this concept, we introduce an incremental skeleton extraction method that generates a lightweight STG representing the observed environment with exploration information and design an FSM to implement STG-based exploration strategy, enabling highly efficient exploration with reduced computational overhead.

\section{METHOD}\label{sec:method}

\begin{figure}[t]
    \centering
    \includegraphics[width=0.45\textwidth]{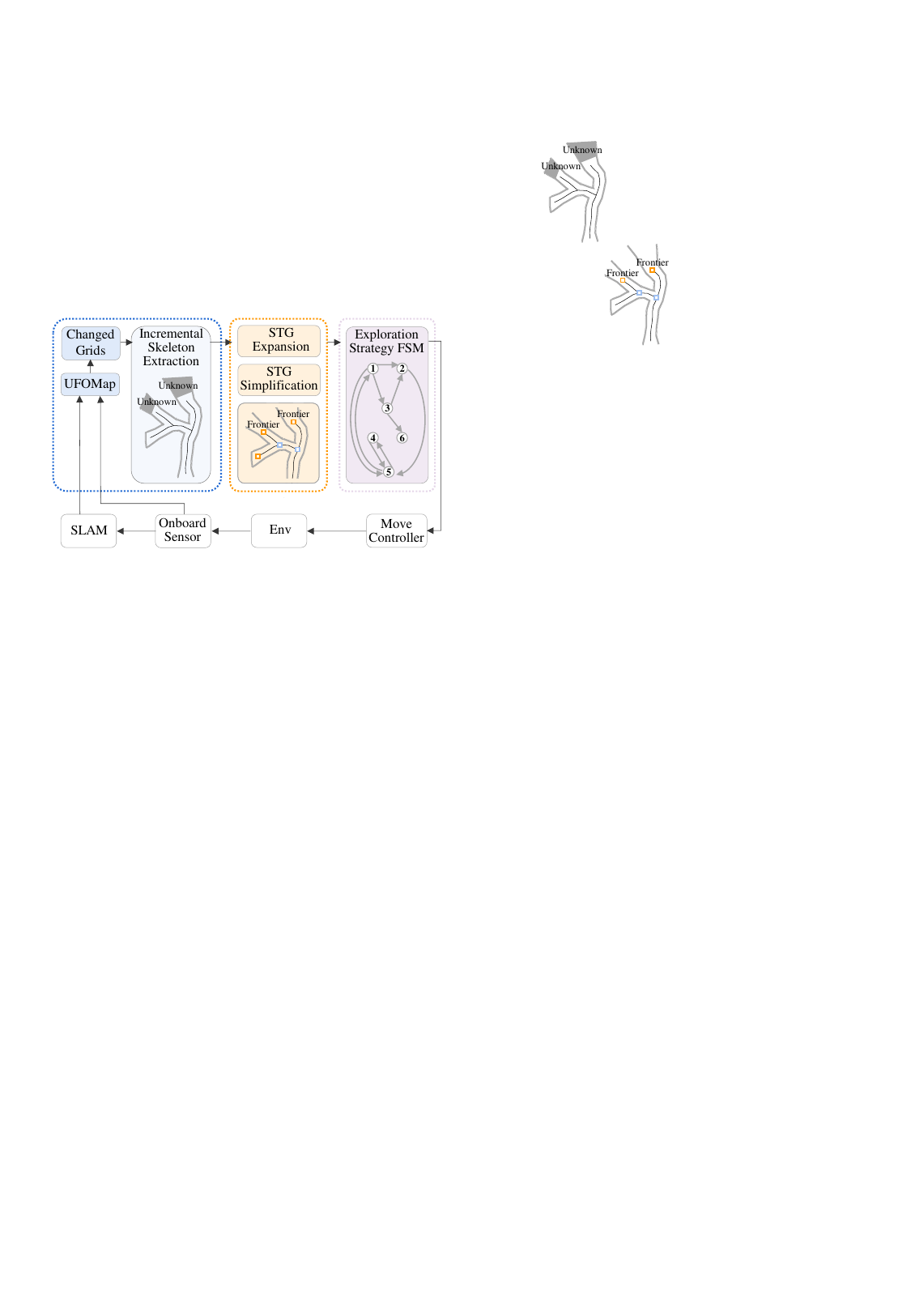}
    \caption{An overview of the proposed STGPlanner.}
    \label{fig:system}
\end{figure}

\subsection{System Overview} 
An overview of the proposed system framework is illustrated in \fref{fig:system}, which consists of three primary modules: incremental skeleton extraction, STG update and an exploration strategy FSM.
The Lidar SLAM module provides localization and registered point clouds, which are used by the 3D occupancy grid map UFOMap \cite{duberg2020UFOMap} to efficiently manage and record environment changes. 
Skeleton extraction (\sref{sec:skeleton}) is performed incrementally using changed grids, allowing for fast updates. 
Based on the current exploration branch and updated skeleton, the STG expands and simplifies (\sref{sec:graph}). 
The FSM then executes the exploration strategy by transitioning states according to the information of STG (\sref{sec:fsm}). 
This cycle repeats until the FSM reaches its termination state.

\subsection{Incremental Skeleton Extraction}\label{sec:skeleton}
%

%

\begin{figure*}[t]
    \centering
    \includegraphics[width=1\textwidth]{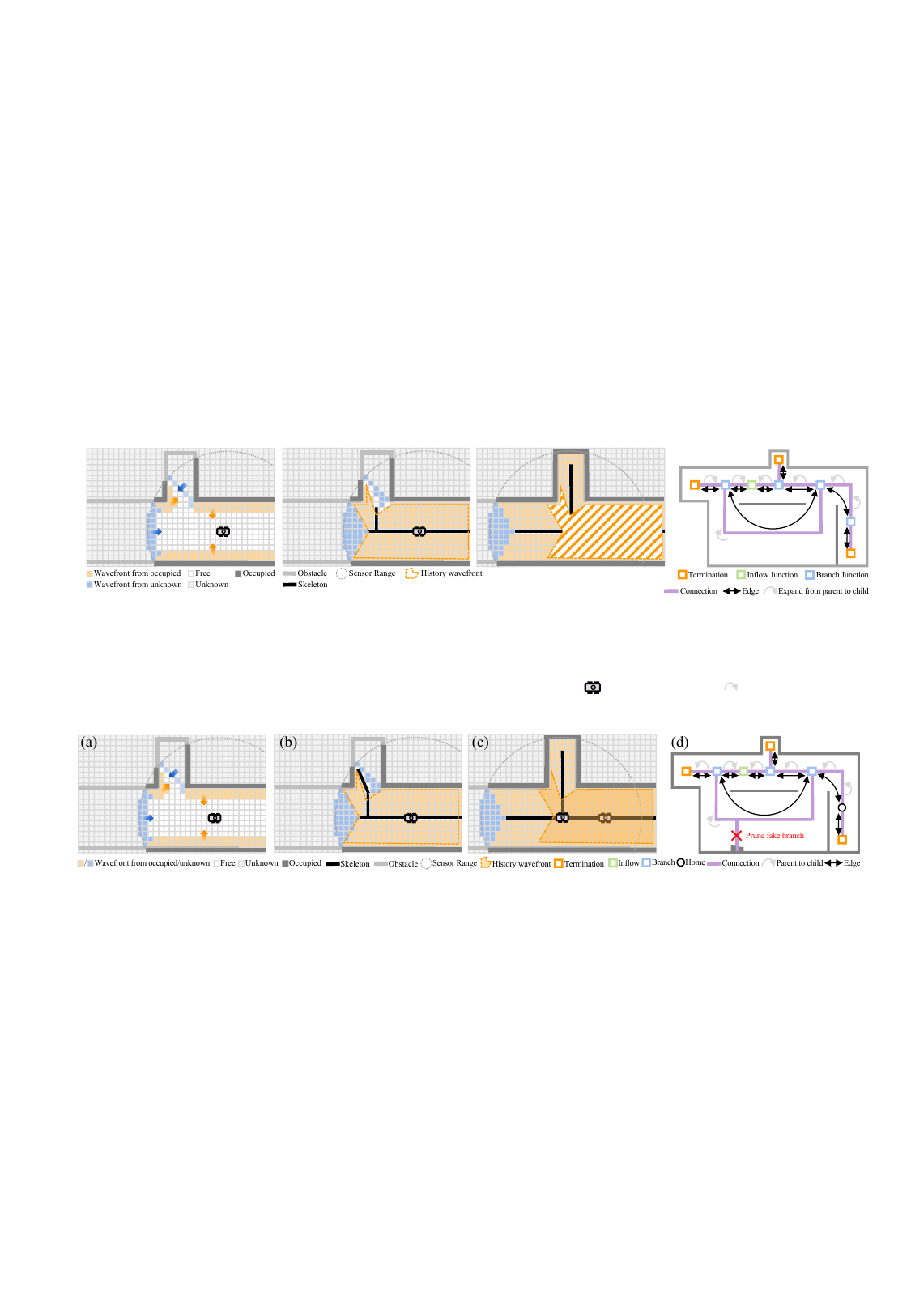}
    \caption{Illustration of the skeleton extraction and STG update. (a) and (b) depict the wavefront propagation from unknown and occupied area at time $T_i$. As robot moves from $T_i$ to $T_{i+1}$ in (c), wavefront from occupied is preserved (shaded), thereby requiring only partial data processing. (d) provides an overview of the STG.}
    \label{fig:stg}
\end{figure*}

To reduce BFMs in exploration, we aim to leverage the topological structure of the observed environment to guide the robot's exploration behavior, which necessitates rapid extraction of a concise topological representation.
The Generalized Voronoi Diagram (GVD) \cite{choset1995Sensora}, which maximizes the distance between obstacles, provides a compact representation of the environment’s geometry.
Given that obstacles are represented by $n$ convex point sets $C_i \subset \mathbb{R}^2$ (where $i = 1, 2, \dots, n$), the two-equidistant face $\mathcal{F}_{ij}$ between sets $C_i$ and $C_j$ is defined as follows:
\begin{equation}
    \begin{aligned}
        \mathcal{S}_{ij}&=\left\{x~{\in}~\mathbb{R}^2~|~d_i(x)=d_j(x),~\nabla d_i(x){\neq}\nabla d_j(x)\right\}, \\
        \mathcal{F}_{ij}&=\left\{x~{\in}~\mathcal{S}_{ij}~|~\forall k{\neq}i,j:d_i(x){\neq}d_k(x)\right\},
    \end{aligned}
\end{equation}
where $d_i(x)$ is the minimum distance from point $x$ to set $C_i$, and $\nabla d_i(x)$ is the direction from $x$ to its nearest point in $C_i$.
The GVD is the union of all $\mathcal{F}_{ij}$, expressed as:
\begin{equation}
    \begin{aligned}
        \mathcal{F}^2=\cup_{i=1}^{n-1}\cup_{j=i+1}^n\mathcal{F}_{ij}.
    \end{aligned}
\end{equation}

%


%
%

%
However, extracting a strict GVD as per the definition incurs high computational complexity, and such precision is unnecessary for exploration.
Inspired by \cite{zhang1984Fast}, we propose an approximate method 
to iteratively identify skeleton and support incremental updates.
Pseudo-code for incremental skeleton extraction is provided in \aref{skeleton}.
The 8-neighborhoods of \(s(i, j)\), \(Adj_8(s)\), consist of the points surrounding \(s(i, j)\), with \(n_1\) as the top neighbor and \(n_2\) to \(n_8\) indexed clockwise.

\linespread{0.8}
\begin{algorithm}[t]
    \LinesNumbered
    \caption{The Incremental Skeleton Extraction}
    \label{skeleton}
    \KwIn{$Skeleton$, $waveFront$}
    \KwOut{The updated skeleton $Skeleton$.}
    $Skeleton^{\prime} \leftarrow Skeleton$\;
    $changed \leftarrow \text{true}$\;
    $iteration \leftarrow 0$\;
    $phase \leftarrow 1$\;
    \While {$changed = \text{true}$}
    {
        $changed \leftarrow \text{false}$\;
        \For {$i=0$ \KwTo $1$}
        {
            $condition = (Skeleton = \mathrm{free})$ \textbf{and} $waveFront \ge iteration$\;
            \For {all $s \in Skeleton$}
            {
                \If {${state(Skeleton, s) \neq \text{free}}$}
                {
                    \textbf{continue}\;
                }
                \If {($phase = 1$ \textbf{and} $(a) \land (b) \land (c) \land (d)$) \textbf{or} ($phase = 2$ \textbf{and} $(a) \land (b) \land (e) \land (f)$)}
                {
                    $waveFront_{s} \leftarrow iteration$\;
                    $changed \leftarrow \text{true}$\;
                    \If{$\exists n_i \in Adj_8(s)$ such that $state(Skeleton, n_i) = \text{unknown}$}
                    {
                        $state(Skeleton^{\prime}, s) \leftarrow \text{unknown}$\;
                    }
                    \Else
                    {
                        $state(Skeleton^{\prime}, s) \leftarrow \text{occupied}$\;
                    }
                }
            }
            $Skeleton \leftarrow Skeleton^{\prime}$\;
            $iteration \leftarrow iteration + 1$\;
            $phase \leftarrow 3 - phase$\;
        }
    }
    \SetKwFunction{Func}{resetUnknownWaveFront}
    \Func{};\\
    \Return $Skeleton$\;
\end{algorithm}

%


%
We iteratively process all changed free grids in the UFOMap until no grid satisfies the specific conditions.
This process resembles wavefront propagation, where waves originate from free grids adjacent to occupied or unknown regions. The skeleton is formed at the points where two waves collide, as illustrated in \fref{fig:stg}. 
Following \cite{zhang1984Fast}, each iteration is divided into two phases with distinct conditions (lines 11-12, \aref{skeleton}), as described below:
\begin{equation}\label{eq:condition}
    \begin{aligned}
    \begin{array}{cccc}
    {(a)} & {A(s)=1}, & {(b)} & {2\leq B(s)\leq6}, \\
    {(c)} & {n_{1}*n_{3}*n_{5}=0}, & {(d)} & {n_{3}*n_{5}*n_{7}=0}, \\
    {(e)} & {n_{1}*n_{3}*n_{7}=0}, & {(f)} & {n_{1}*n_{5}*n_{7}=0}, \\
    \end{array}
    \end{aligned}
\end{equation}
where $A(s)$ represents the number of transitions from 0 to 1 in the ordered sequence $n_1, n_2, \cdots, n_8$, and $B(s)$ denotes the number of nonzero neighbors of $s$, defined as:
\begin{equation}
    \begin{aligned}
        B(s)=n_{1}+n_{2}+n_{3}+\cdots+n_{7}+n_{8},
    \end{aligned}
\end{equation}
where $n_i=1$ if $state(s_{ni}) = free$ otherwise $n_i=0$.
To support incremental updates, each grid cell that satisfies the specified condition not only updates its state but also records the current iteration as its wavefront. This value is then used to determine the parameter \(n_i\) in \eref{eq:condition} via $n_i = condition_{s_{n_i}}$ (lines 8-13, \aref{skeleton}).
Moreover, unknown regions are treated as special occupied grids; all grids covered by waves triggered from these grids are marked as unknown, and, ultimately, $waveFront$ and $Skeleton$ within these unknown regions are reset (lines 15-18, 22, \aref{skeleton}).
This procedure both avoids redundant computation of obstacle-triggered wavefront propagation and accounts for the uncertainty of unknown areas, thereby ensuring the coherence and stability of skeleton, as shown in \fref{fig:stg}.
Additionally, $waveFront$ provides insight into the path's width and is subsequently applied in Secs.~\ref{sec:simplify} and~\ref{sec:fsm}.

\subsection{Skeletal Topological Graph Update}\label{sec:graph}
The update process of the STG, denoted as \(\mathcal{G} = (\mathcal{N}, \mathcal{E})\), is detailed in \aref{graph}. 
To accurately document the exploration process and facilitate decision-making, the graph is constructed in a tree-growth-inspired manner, with nodes classified into four distinct types:
\begin{itemize}
    \item \textbf{Termination} (\(N_T\)): One parent, no children.
    \item \textbf{Connection} (\(N_{C}\)): One parent, one child.
    \item \textbf{Branch Junction} (\(N_{B}\)):  One parent, multiple children.
    \item \textbf{Inflow  Junction} (\(N_{I}\)): Multiple parents, any children.
\end{itemize}

\subsubsection{Expansion}
During initialization, the skeleton grid closest to the starting point is marked as $s_{Home}$.
\textbf{genNode($s_{Home}$)} is called to generate the corresponding node \(n_{Home} \in \mathcal{N}\), which is added to the frontier candidates queue \(\mathcal{N}_f\). 
During exploration, for each node in \(\mathcal{N}_f\), its 8-neighborhoods is traversed to generate nodes associated with free grids in $Skeleton$, and neighbor nodes are designated as its child nodes (lines 15-19, \aref{graph}).
Nodes adjacent to unknown grids are added to \(\mathcal{N}_f\), serving as starting points for the next expansion phase (line 5, \aref{graph}).
If a neighboring node $n_{s_{ni}}$ already exists, the pair \(\{(n_{s_{ni}}, n_s)\}\) is added to the inflow junction candidates queue \(\mathcal{N}^2_i\) (line 8, \aref{graph}).
Subsequently, \textbf{setNodeType($n$)} is invoked to classify the node according to predefined criteria.
All newly generated child nodes recursively follow this process to achieve depth-first graph expansion (lines 13-14, \aref{graph}).

\linespread{0.8}
\begin{algorithm}[t]
    \LinesNumbered
    \caption{The Graph Generation}
    \label{graph}
    \KwIn{$Skeleton$, $waveFront$, $\mathcal{G}$, $\mathcal{N}_f$}
    \KwOut{The updated graph $\mathcal{G}$ and frontier set $\mathcal{N}_f$.}
    
    \SetKwFunction{FuncA}{genNode} 
    \SetKwFunction{FuncB}{setNodeType}
    \SetKwFunction{FuncC}{expandRec}
    
    
    \SetKwProg{Fn}{Function}{:}{end} 
    
    \Fn{\FuncC{$n_s$}}{
        $childQueue \gets \texttt{empty}$\;
        \For {all $s_{ni}$ in $Adj_8(s)$}
        {
            \If {$ state(skeleton,s_{ni}) = unknown$}
            {
                $\mathcal{N}_f \gets \mathcal{N}_f \cup n_s$\;
            }
            \ElseIf {$ state(skeleton,s_{ni}) = free$}
            {
                \If {$n_{s_{ni}} \in \mathcal{N}$}
                {
                   $\mathcal{N}^2_i \gets \mathcal{N}^2_i \cup \{(n_{s_{ni}}, n_s)\}$;\
                }
                \Else
                {
                    \FuncA{$s_{ni}$};\\
                    $childQueue \gets childQueue \cup n_{s_{ni}}$\;
                }
            }
        }
        \FuncB{$n_s$};\\
        \For {all $n'_s$ in $childQueue$}
        {
            \FuncC{$n'_s$};\
        }
    }

    \SetKwFunction{FuncD}{procInflow}
    \SetKwFunction{FuncE}{pruneGraph}

    %
    
    \SetKwFunction{FuncF}{updateEdges}

        
    
    \tcc{start main function}
    $\mathcal{N'}_f \leftarrow \mathcal{N}_f$\;
    $\mathcal{N}_f \gets \texttt{empty}$\;
    $\mathcal{N}^2_i \gets \texttt{empty}$\;
    \For {all $n_s$ in $\mathcal{N'}_f$}
    {
        \FuncC{$n_s$};\
    }
    \FuncD{$\mathcal{N}^2_i$, $step$};\\
    \FuncE{$\mathcal{G}$, $waveFront$, $thres$};\\
    \FuncF{$\mathcal{G}$};\\
    \Return $\mathcal{G}$, $\mathcal{N}_f$\;
\end{algorithm}

\subsubsection{Simplification}\label{sec:simplify}
We process inflow junction candidates first.
Each pair in $\mathcal{N}^2_i$ is traced back towards their respective parent nodes within a specified step range $step$. 
If a common parent is found within $step$, the pair is identified as a fake inflow and no edge is established. Otherwise, one of the pair is designated as $N_I$, and the information of the relevant nodes is updated accordingly.

The uneven contours of obstacles can lead to artifacts or spurious features during skeleton extraction, potentially distorting the environment's topological structure, as shown in \fref{fig:stg}d.
When two waves converge with a short propagation distance, it may indicate irregular obstacle shapes or sensor errors.
Based on this assumption, if the wavefront value of newly generated \(N_T\) nodes is below the threshold $thres$, the entire branch is traced back and pruned until a \(N_B\) or \(N_C\) node is encountered.

Edges are established between adjacent nodes in a parent-child hierarchy, while \(N_C\) nodes are omitted to simplify \(\mathcal{G}\). 
Consequently, edges exist only between \(N_B\), \(N_C\), and \(N_T\) nodes, as illustrated in \fref{fig:stg}d. 
The sequence and length of the omitted \(N_C\) nodes are recorded as attributes of the edges, which are later used for path searching.

\subsection{STG-based exploration strategy}\label{sec:fsm} 
\begin{figure}[t]
    \centering
    \includegraphics[width=0.4\textwidth]{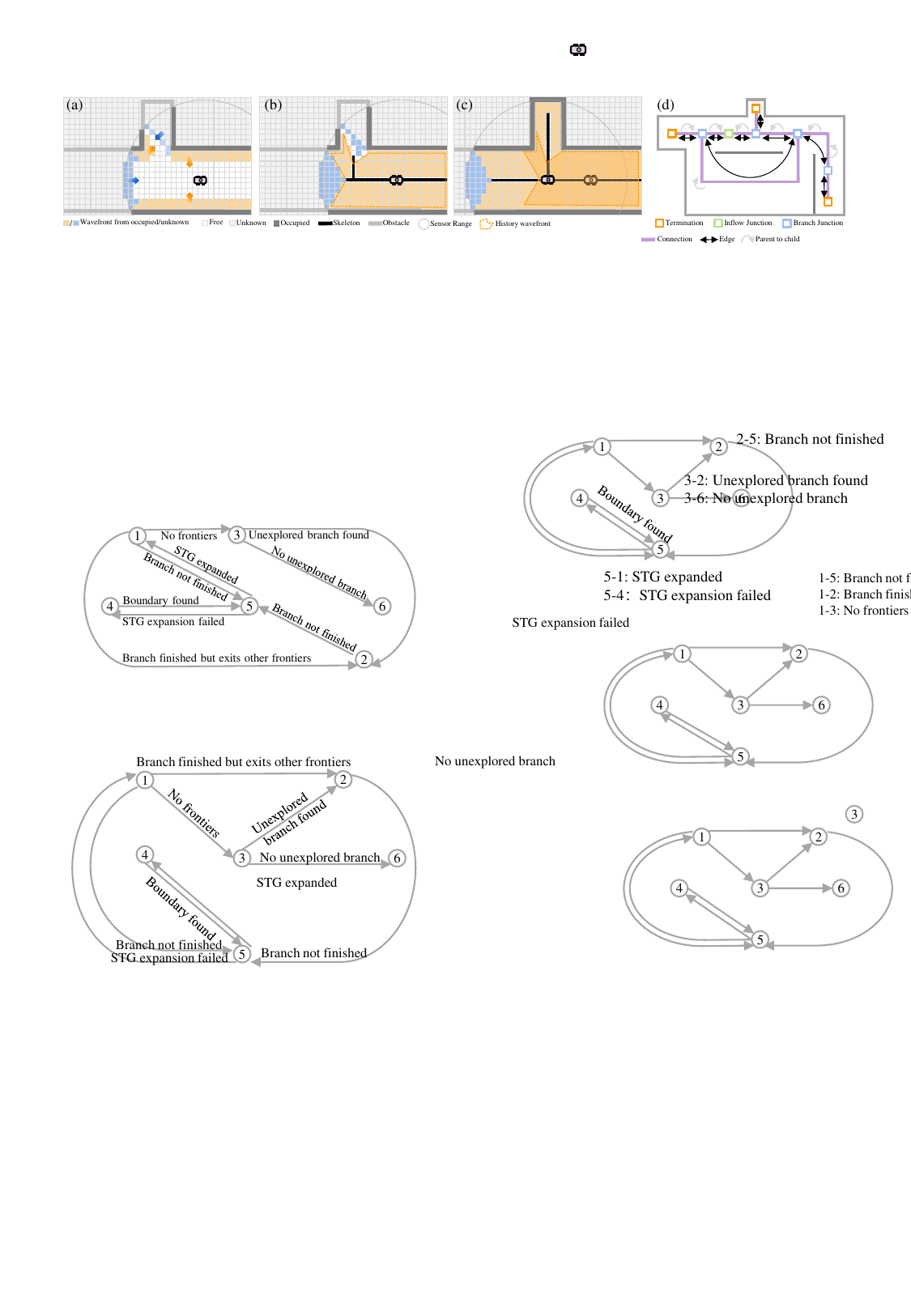}
    \caption{Illustration of the exploration strategy FSM.}
    \label{fig:fsm}
\end{figure}

The STG inherently encodes information about frontiers, viewpoints, gains, and historical trajectories. 
Leveraging this, we propose a computationally efficient exploration strategy that eliminates the need for ray-casting and TSP-solving.
This strategy is implemented through a FSM, as illustrated in \fref{fig:fsm}, which comprises six distinct states.

\textbf{State 1: Topology-Guided Exploration.}
This is the primary state of the exploration strategy.
If $\mathcal{N}_f$ is not empty, classify all frontiers into branches based on their parent junction ($N_I$ or $N_B$). If the current branch contains frontiers, transition to \textbf{State 5} with the nearest frontier to prioritize the completion of current branch, which effectively reducing BFMs. Otherwise, move to \textbf{State 2} to update the exploration branch. If $\mathcal{N}_f$ is empty, the exploration of the current region is complete, and the system transitions to \textbf{State 3} for backtracking.
Moreover, STG positions targets centrally within paths, enhancing observation and navigation safety.

\textbf{State 2: Exploration Branch Update.}
The exploration branch is updated once the current one is completed.
Following the small-region-first approach outlined in \cite{wei2022Robot}, we greedily select the branch with the smallest wavefront value as the new exploration branch, as a smaller wavefront indicates a narrower region.
Unselected branches are grouped and stored in the $missedBranches$ stack.
The system then transitions to \textbf{State 5} with the nearest frontier in the updated exploration branch.

\textbf{State 3: Backtracking.}
In this state, the system continuously pops entries from $missedBranches$ in chronological order to locate unexplored branches. Once a valid group is found, it transitions to \textbf{State 2} to update the exploration branch. If the stack is empty, it indicates that the environment has been fully explored, and the system transitions to the termination state, \textbf{State 6}.

\textbf{State 4: Boundary-Guided Exploration.}
In overly open environments, the sensor's perception range limits the expansion of the STG, hindering the generation of effective frontier nodes. In this case, this state is activated to perform traditional boundary exploration.
Following \cite{huang2023FAEL}, the boundary between known and unknown regions is detected 
. A 2-opt heuristic algorithm is then applied to plan the path, transitioning to \textbf{State 5} with the first grid.

\textbf{State 5: Planning and Moving.}
In this state, A$^*$ is employed for path planning based on the target \(s\).
If \(s\) is close to the agent and \(state(skeleton,s)=unknown\), it indicates that the STG has not effectively expanded.
Therefore, the system will transition to \textbf{State 4} to replan the target. Otherwise, the system will invoke the motion controller to proceed to the target and return to \textbf{State 1}.
Furthermore, due to the structure of the STG, the backtracking phase allows for the rapid extraction of detailed paths from edges, significantly reducing the search space of A$^*$.

\textbf{State 6: Exploration Termination.}

\section{EXPERIMENTAL RESULTS}\label{sec:result}

\begin{figure}[t]
    \centering
    \subfloat[Scene1: 150m$\times$110m]{
        \includegraphics[width=0.47\linewidth]{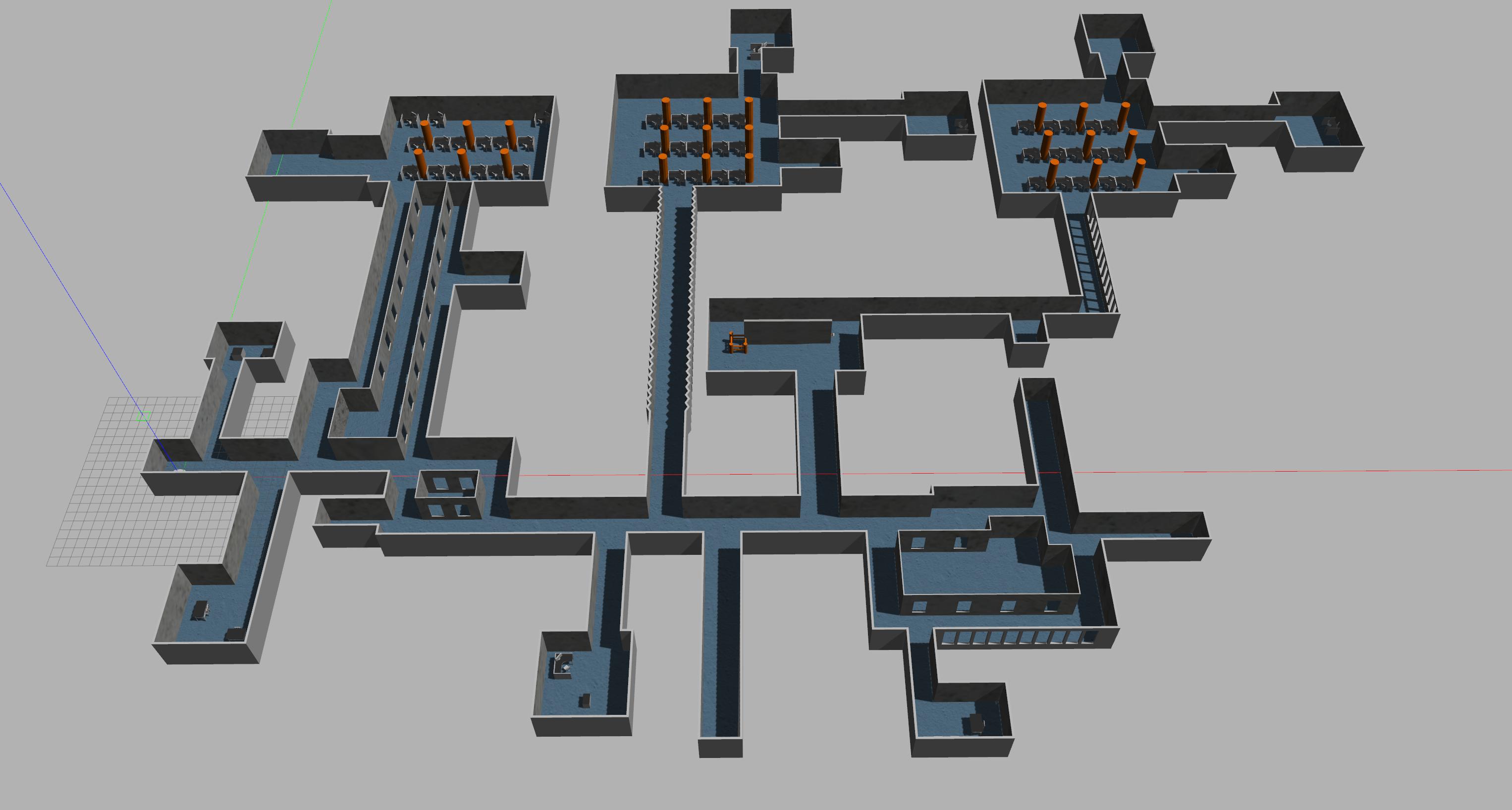}}
    \subfloat[Scene2: 100m$\times$90m]{
        \includegraphics[width=0.47\linewidth]{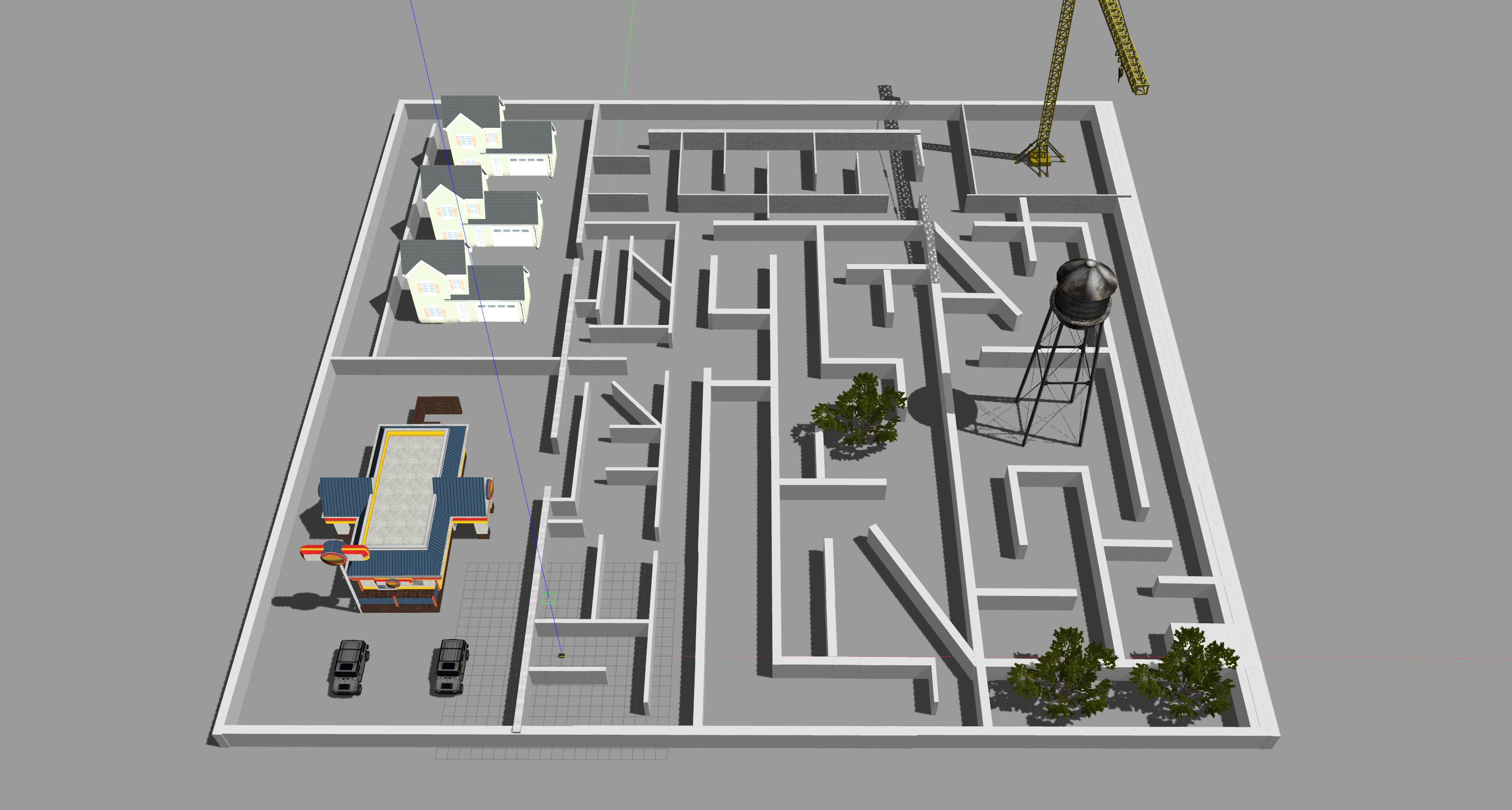}}
        
    \subfloat[Scene3: 210m$\times$155m]{
        \includegraphics[width=0.47\linewidth]{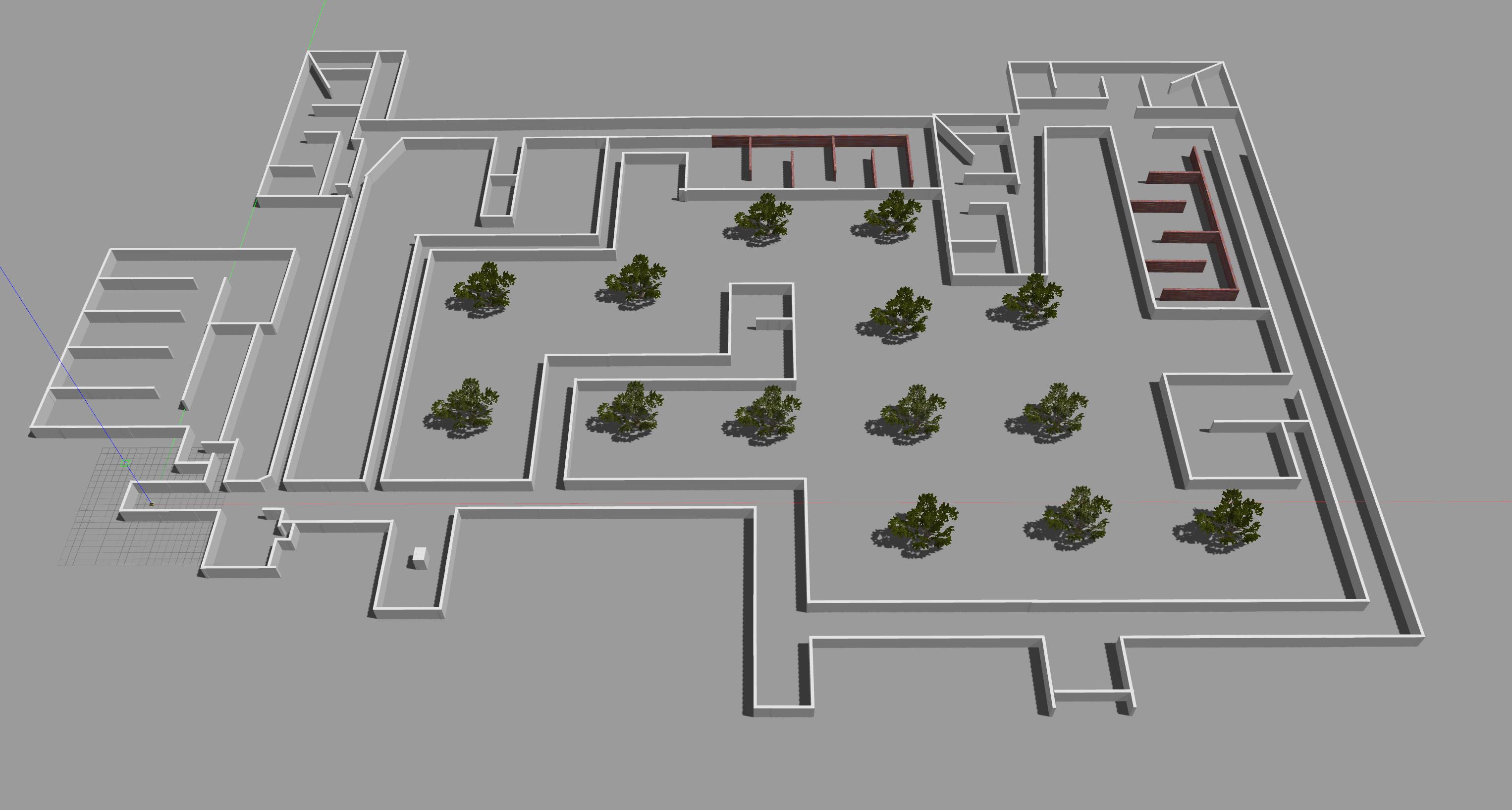}}
    \subfloat[Scene4: 130m$\times$120m]{
        \includegraphics[width=0.47\linewidth]{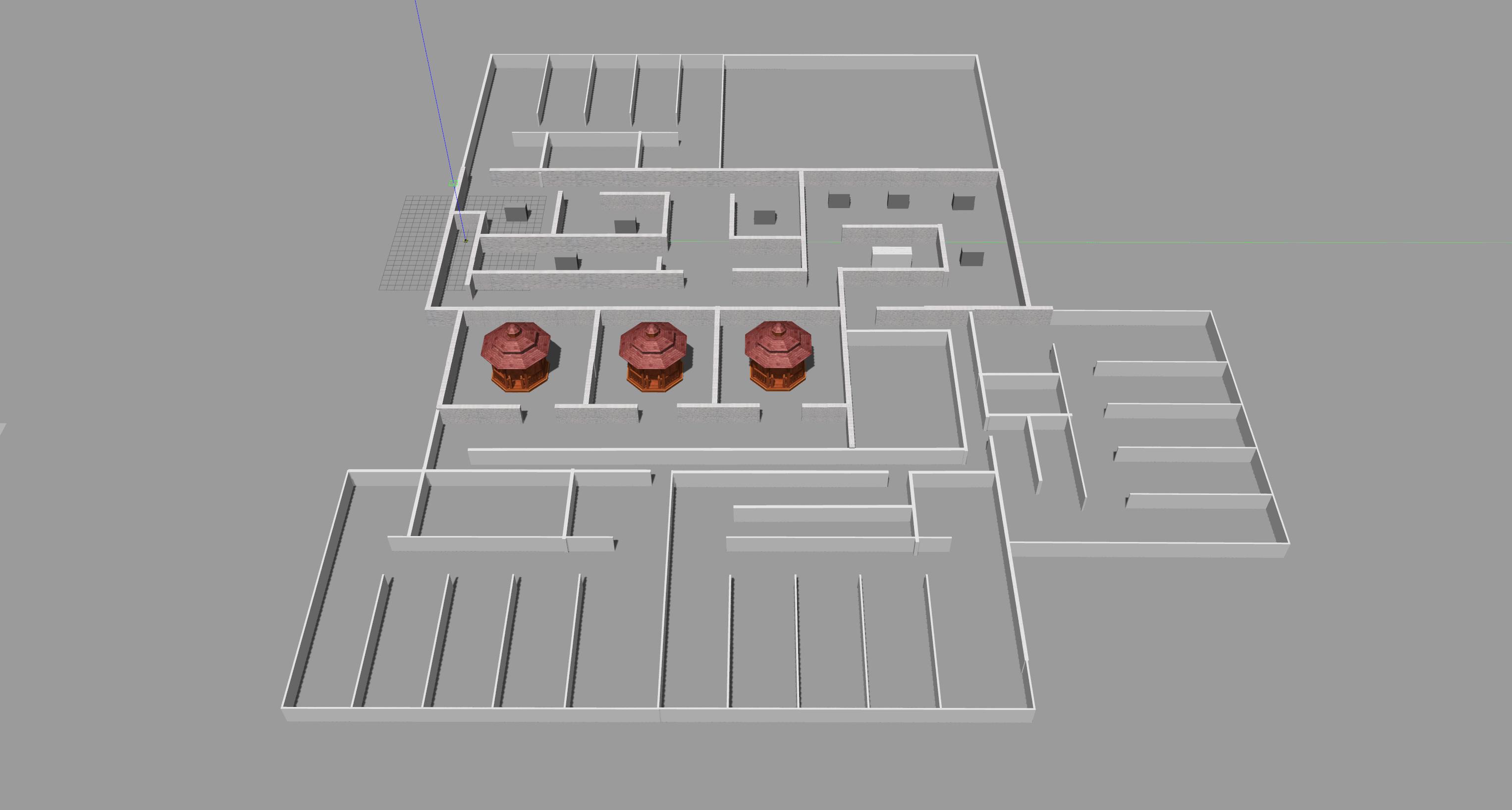}}
    
    \caption{The four simulation environments.}
    \label{fig:sim_env}
\end{figure}

\subsection{Benchmark Comparison}\label{sec:Benchmark}
We conduct a series of experiments to evaluate the proposed method in various environments, as illustrated in \fref{fig:sim_env}.
The comparative methods include sampling-based methods GBP2\cite{dang2020Graphbased} and TARE\cite{cao2021TARE}, as well as the frontier-based method FAEL\cite{huang2023FAEL}, the latter two of which incorporates different forms of TSP solving.
All methods are implemented using open-source code. 
To ensure a fair comparison of the exploration strategies, we standard the motion planner by using a modified version of \cite{zhang2020Falco} with  $v_{max}=2.0 m/s$.
We employ an open-source ground robot simulator \footnote{\url{https://github.com/jackal/jackal}}, equipped with a Velodyne VLP-16 LiDAR with a FoV [360°, 30°], as the simulation platform. 
Each method is tested five times under identical configurations on a computer equipped with a 3.70 GHz Intel i9-10900K CPU running Ubuntu 20.04.

\subsubsection{Exploration Performance}\label{sec:synthia_setup}

\begin{figure}[t]
    \centering
    
    \subfloat[Scene1]{
        \label{subfig:Scene1}
        \includegraphics[width=0.47\linewidth]{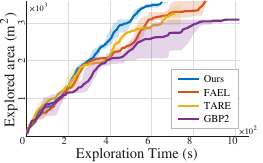}}
    \subfloat[Scene2]{
        \label{subfig:Scene2}
        \includegraphics[width=0.47\linewidth]{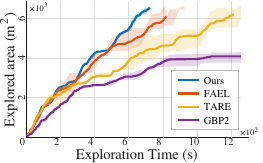}}
        
        
    \subfloat[Scene3]{
        \label{subfig:Scene3}
        \includegraphics[width=0.47\linewidth]{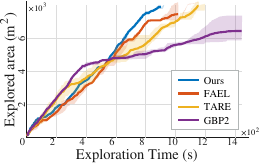}}
    \subfloat[Scene4]{
        \label{subfig:Scene4}
        \includegraphics[width=0.47\linewidth]{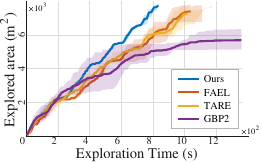}}
    
    \caption{The explored area over time. The shaded areas show the upper and lower bounds.}
    \label{fig:sim_result}
\end{figure}

The quantitative results are shown in \tref{tab:simulation}, 
where bold represents the best.
It is evident that our method demonstrates superior exploration performance in all four scenes, reducing exploration time by at least 19.8\% and path length by at least 15.6\%.
To offer a more intuitive understanding of the performance, we also plot the curve of explored area over time, as shown in \fref{fig:sim_result}.
In the initial phase, all methods demonstrate comparable performance, and our method occasionally underperforms, as observed in scene 1 and scene 3. 
However, as exploration continues, the advantages of our method become progressively more apparent.

In the experiments of TARE and FAEL, the robot occasionally exhibited behaviors such as oscillation in place or BFMs at junctions and in narrow passages, which significantly impedes exploration efficiency.
These behaviors stem from unreliable evaluation of viewpoint information gain under incomplete observations, which leads to instability in the 'optimal' trajectory solved by TSP.
As the robot gathers new information during movement, the current optimal trajectory may conflict with the previously determined ones, especially in complex scenarios.
This issue can also be observable in \fref{fig:sim_result}, where TARE and FAEL display more pronounced stepped profiles. 
This stepping pattern is due to leaving areas before they are fully explored, which necessitates frequent backtracking.

However, while our method, guided by the environmental topological structure, may not always achieve the maximum information gain at each stage, it ensures comprehensive exploration of the current branch in a single pass without inefficient BFMs. 
This method ultimately provides greater long-term benefits and results in improved exploration performance.

\subsubsection{Computational Performance}\label{sec:compute}

\begin{table}[t]
\aboverulesep=-1pt
\belowrulesep=0pt %
 \caption{Comparison with SOTA methods.}
   \label{tab:simulation}
\centering
\scalebox{0.85}{
\renewcommand{\arraystretch}{1.0}  %
\begin{tabular}{c|c|cc|cc|c|c} 
\toprule[2pt]
\multirow{2}{*}{\textbf{Scene}}  
    & \multirow{2}{*}{\textbf{Method}} 
    & \multicolumn{2}{l|}{\makecell{\textbf{Exploration}\\ \textbf{time(s)}}} 
    & \multicolumn{2}{l|}{\makecell{\textbf{Path}\\ \textbf{length(m)}}} 
    & \multirow{2}{*}{\makecell{\textbf{Run}\\ \textbf{time(s)}}}
    & \multirow{2}{*}{\makecell{\textbf{Speed}\\ \textbf{(m/s)}}} \\ 
\cline{3-6}
& & {\textbf{Avg}} & \textbf{Std} 
    & \textbf{Avg} & \textbf{Std} 
    & & \\ 
\specialrule{1.5pt}{-1pt}{1pt} %

\multirow{4}{*}{Scene1}
    & GBP2\cite{dang2020Graphbased}
        & \textgreater1200 & - & \textgreater954.4 & - & 0.337 & 0.80\\
    & TARE\cite{cao2021TARE}
        & 733.8 & 54.9 & 1118.9 & 77.2 & 0.299 & 1.52 \\
    & FAEL\cite{huang2023FAEL}
        & 786.4 & 92.0 & 1075.8  & 86.8 & 0.041 & 1.37 \\
    & Ours 
        & \textbf{541.6} & 18.4 & \textbf{875.9} & 33.5 & \textbf{0.020} & \textbf{1.62} \\
\specialrule{0.5pt}{-0.5pt}{1pt}

\multirow{4}{*}{Scene2}
    & GBP2\cite{dang2020Graphbased}
        & \textgreater1500 & - & \textgreater1411.1 & - & 0.795 & 0.95 \\
    & TARE\cite{cao2021TARE}
        & 1181.4 & 102.5 & 1716.6 & 21.7 & 0.375 & 1.45 \\
    & FAEL\cite{huang2023FAEL}
        & 883.1 & 52.6 & 1266.8  & 38.6 & 0.082 & 1.52 \\
    & Ours 
        & \textbf{681.4} & 22.1 & \textbf{1068.9} & 3.8 & \textbf{0.008} & \textbf{1.57} \\
\specialrule{0.5pt}{-0.5pt}{1pt}

\multirow{4}{*}{Scene3}
    & GBP2\cite{dang2020Graphbased}
        & \textgreater1700 & - & \textgreater1915.8 & - & 0.377 & 1.12 \\
    & TARE\cite{cao2021TARE}
        & 1104.7 & 29.2 & 1889.4 & 44.8 & 0.312 & \textbf{1.71} \\
    & FAEL\cite{huang2023FAEL}
        & 1020.7 & 59.7 & 1674.7  & 126.6 & 0.050 & 1.64 \\
    & Ours 
        & \textbf{818.1} & 21.2 & \textbf{1372.4} & 24.8 & \textbf{0.033} & 1.67 \\
\specialrule{0.5pt}{-0.5pt}{1pt}

\multirow{4}{*}{Scene4}
    & GBP2\cite{dang2020Graphbased}
        & \textgreater1600 & - & \textgreater1560 & - & 0.465 & 0.97 \\
    & TARE\cite{cao2021TARE}
        & 1018.6 & 82.3 & 1640.0 & 135.3 & 0.338 & \textbf{1.61} \\
    & FAEL\cite{huang2023FAEL}
        & 1033.9 & 149.8 & 1420.3  & 128.4 & 0.119 & 1.37 \\
    & Ours 
        & \textbf{779.8} & 17.2 & \textbf{1194.0} & 12.9 & \textbf{0.015} & 1.53 \\

\bottomrule[2pt]
\end{tabular}}
\end{table}

\tref{tab:simulation} shows that our runtime significantly outperforms the compared methods.
On average, STGPlanner demonstrates performance 4+ times faster than FAEL, 22+ times faster than TARE, and 38+ times faster than GBP2.
The high runtime of these comparison methods primarily arises from three factors: solving the TSP, constructing RRTs, ray-casting-based gain evaluations, as noted in \cite{tang2023Bubble}. 
Additionally, in environments with narrow and branching structures, reducing the sampling step size to ensure comprehensive exploration further exacerbates TSP complexity and increases the frequency of ray-casting operations.
In contrast, our method uses a lightweight STG to retain environmental information and employs a computationally efficient FSM to make exploration decisions based on the topological structure.
Additionally, the tree-like growth structure of the STG significantly reduces the path search space during long-distance backtracking.
These result in a significantly lower runtime throughout the exploration process.

\begin{figure}[t]
    \centering
    \subfloat[Runtime of Scene1.]{
        \label{fig:runtime}
        \includegraphics[width=0.47\linewidth]{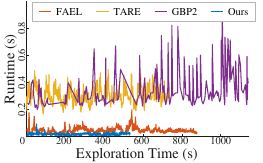}}
    \subfloat[Graph size of Scene1.]{
        \label{fig:graphsize}
        \includegraphics[width=0.47\linewidth]{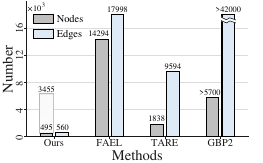}}

    \caption{Illustration of computational performance.}
    \label{fig:computational performance}
\end{figure}

We further record and plot the runtime during exploration in Scene1, as shown in \fref{fig:runtime}.
GBP2 re-plans only after reaching last target, resulting in a low decision frequency. In later stages of exploration, it often oscillates between distant points with high computational overhead.
In contrast, TARE and FAEL make decisions in real-time. 
The increase in candidate viewpoints in certain areas, such as junctions or expansive regions, significantly escalates the time required to solve the TSP. 
This contributes to the runtime spikes observed in \fref{fig:runtime}.
Our method, on the other hand, maintains a consistently low runtime, with skeleton extraction requires approximately 12.3 ms, while STG updates and FSM decision-making take around 7.8 ms.

%
Moreover, We quantify the size of the graph generated by each method after completing the exploration in Scene1, as depicted in \fref{fig:graphsize}. In our method, the gray bar represents  \(N_B\), \(N_I\), and \(N_T\) nodes, while \(N_C\) nodes are shown separately as blank bar, since they are removed and do not contribute to edge construction.
Compared to methods that generate roadmap through sampling, our proposed STG, which captures environmental topological characteristics, is significantly smaller in size. This reduction not only enhances exploration efficiency but also supports navigation tasks more effectively.


\subsection{Real-World Experiment}

\begin{figure}[t]
    \centering
    
    \subfloat[The indoor environment to be explored.]{
        \label{subfig:real_env}
        \includegraphics[width=0.9\linewidth]{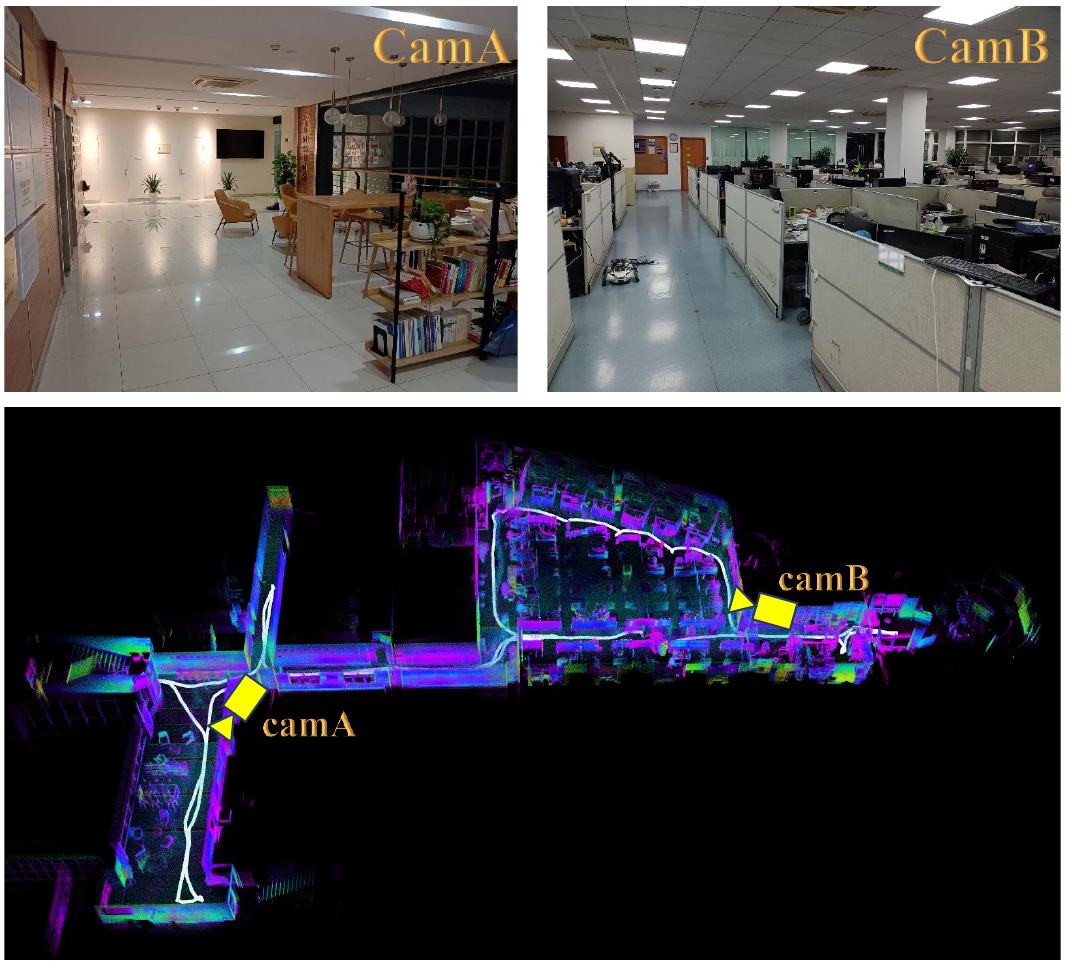}}
    
    \subfloat[The platform.]{
        \label{subfig:platform}
        \includegraphics[width=0.3\linewidth]{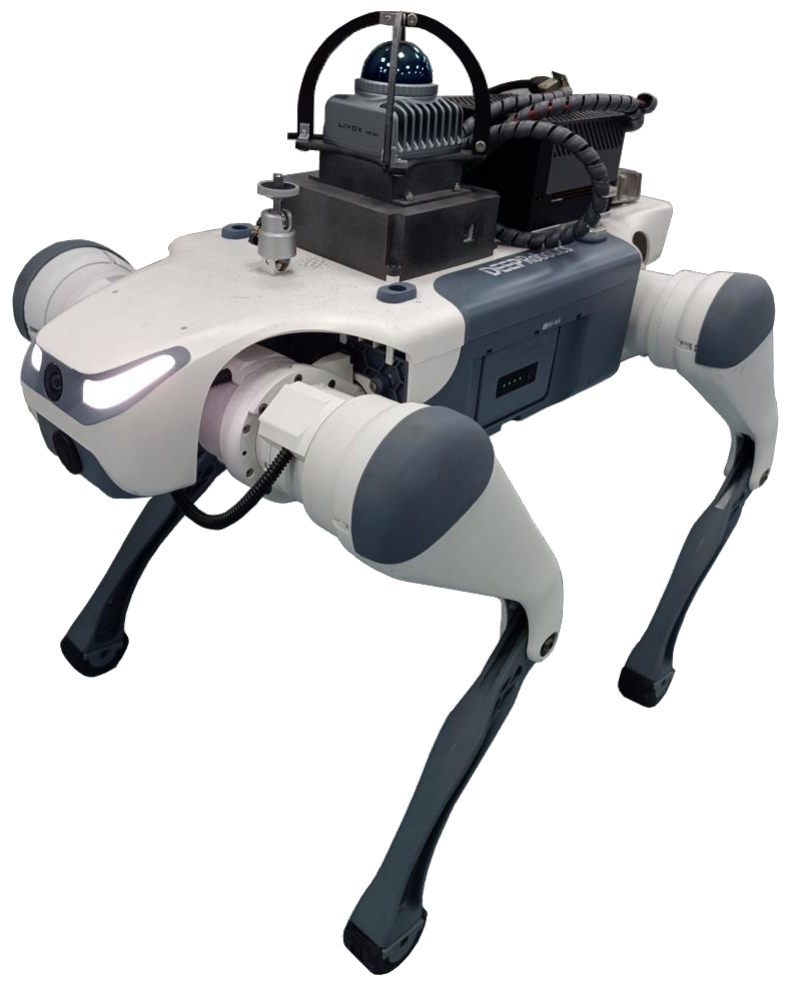}}
    \subfloat[The explored area over time.]{
        \label{subfig:real_result}
        \includegraphics[width=0.6\linewidth]{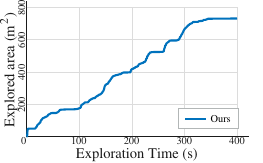}}
    
    \caption{Exploration experiment conducted in real-world.}
    \label{fig:airsimPR}
\end{figure}
In an indoor office environment, we deployed a quadrupedal robot equipped with a Jetson AGX Orin and Livox Mid-360 LiDAR, using FAST-LIO2\cite{zhang1984Fast} for real-time localization.
\fref{subfig:real_env} shows the generated point‐cloud map and exploration trajectory (white line).
The exploration of approximately 731$m^2$ is completed in about 350s, with a total travel distance of 171m.
This indicates that our proposed framework is able to perform environment exploration tasks on physical platforms with both safety and efficiency.

\section{CONCLUSION}\label{sec:conclusion}
In this paper, we introduced STGPlanner, a novel autonomous exploration framework based on skeletal topological graph (STG). 
We proposed an incremental skeleton extraction method that allows the STG to expand effectively. Further, a FSM was designed to encapsulate the entire exploration strategy. 
By leveraging the topological structure of the environment, our method significantly reduces BFMs, enhancing exploration efficiency while minimizing computational costs. 
However, several limitations remain. 
In larger open environments, skeletal extraction may prove ineffective, necessitating a reliance on traditional frontier-based methods. Furthermore, sensor noise and dynamic elements in real-world scenarios can adversely affect the skeletal structure.
Future work will focus on adapting the system for dynamic environments and facilitating multi-robot cooperation.











\bibliographystyle{IEEEtran}
\IEEEtriggeratref{13}
\bibliography{reference}

@ARTICLE{zhao2024Autonomous,
  author={Zhao, Yinghao and Yan, Li and Xie, Hong and Dai, Jicheng and Wei, Pengcheng},
  journal={IEEE Transactions on Industrial Electronics}, 
  title={Autonomous Exploration Method for Fast Unknown Environment Mapping by Using UAV Equipped With Limited FOV Sensor}, 
  year={2024},
  volume={71},
  number={5},
  pages={4933-4943},
  doi={10.1109/TIE.2023.3285921}}

@ARTICLE{hui2024DPPM,
  author={Hui, Yulin and Zhang, Xuewei and Shen, Hongming and Lu, Hanchen and Tian, Bailing},
  journal={IEEE Transactions on Intelligent Vehicles}, 
  title={DPPM: Decentralized Exploration Planning for Multi-UAV Systems Using Lightweight Information Structure}, 
  year={2024},
  volume={9},
  number={1},
  pages={613-625},
  doi={10.1109/TIV.2023.3322705}}

@ARTICLE{luo2024StarSearcher,
  author={Luo, Yiming and Zhuang, Zixuan and Pan, Neng and Feng, Chen and Shen, Shaojie and Gao, Fei and Cheng, Hui and Zhou, Boyu},
  journal={IEEE Robotics and Automation Letters}, 
  title={Star-Searcher: A Complete and Efficient Aerial System for Autonomous Target Search in Complex Unknown Environments}, 
  year={2024},
  volume={9},
  number={5},
  pages={4329-4336},
  doi={10.1109/LRA.2024.3379840}}

@ARTICLE{lindqvist2024TreeBased,
  author={Lindqvist, Björn and Patel, Akash and Löfgren, Kalle and Nikolakopoulos, George},
  journal={IEEE Transactions on Robotics}, 
  title={A Tree-Based Next-Best-Trajectory Method for 3-D UAV Exploration}, 
  year={2024},
  volume={40},
  number={},
  pages={3496-3513},
  doi={10.1109/TRO.2024.3422052}}

@article{chen2023STExplorer,
author = {Chen, Bolei and Cui, Yongzheng and Zhong, Ping and Yang, Wang and Liang, Yixiong and Wang, Jianxin},
title = {STExplorer: A Hierarchical Autonomous Exploration Strategy with Spatio-temporal Awareness for Aerial Robots},
year = {2023},
issue_date = {December 2023},
publisher = {Association for Computing Machinery},
address = {New York, NY, USA},
volume = {14},
number = {6},
issn = {2157-6904},
doi = {10.1145/3595184},
journal = {ACM Transactions on Internet Technology},
month = {Nov.},
articleno = {99},
numpages = {24}
}

@INPROCEEDINGS{tang2023Bubble,
  author={Tang, Benxu and Ren, Yunfan and Zhu, Fangcheng and He, Rui and Liang, Siqi and Kong, Fanze and Zhang, Fu},
  booktitle={2023 IEEE/RSJ International Conference on Intelligent Robots and Systems (IROS)}, 
  title={Bubble Explorer: Fast UAV Exploration in Large-Scale and Cluttered 3D-Environments Using Occlusion-Free Spheres}, 
  year={2023},
  volume={},
  number={},
  pages={1118-1125},
  doi={10.1109/IROS55552.2023.10342348}}

@INPROCEEDINGS{xu2023Heuristicbased,
  author={Xu, Zhefan and Suzuki, Christopher and Zhan, Xiaoyang and Shimada, Kenji},
  booktitle={2024 IEEE International Conference on Robotics and Automation (ICRA)}, 
  title={Heuristic-based Incremental Probabilistic Roadmap for Efficient UAV Exploration in Dynamic Environments}, 
  year={2024},
  volume={},
  number={},
  pages={11832-11838},
  doi={10.1109/ICRA57147.2024.10610462}}

@INPROCEEDINGS{kim2023Topological,
  author={Kim, Boseong and Seong, Hyunki and Shim, D. Hyunchul},
  booktitle={2024 IEEE International Conference on Robotics and Automation (ICRA)}, 
  title={Topological Exploration using Segmented Map with Keyframe Contribution in Subterranean Environments}, 
  year={2024},
  volume={},
  number={},
  pages={6199-6205},
  doi={10.1109/ICRA57147.2024.10610605}}

@INPROCEEDINGS{zhao2023TDLE,
  author={Zhao, Xuyang and Yu, Chengpu and Xu, Erpei and Liu, Yixuan},
  booktitle={2023 IEEE 19th International Conference on Automation Science and Engineering (CASE)}, 
  title={TDLE: 2-D LiDAR Exploration with Hierarchical Planning Using Regional Division}, 
  year={2023},
  volume={},
  number={},
  pages={1-6},
  doi={10.1109/CASE56687.2023.10260441}}

@ARTICLE{zhou2023RACER,
  author={Zhou, Boyu and Xu, Hao and Shen, Shaojie},
  journal={IEEE Transactions on Robotics}, 
  title={RACER: Rapid Collaborative Exploration With a Decentralized Multi-UAV System}, 
  year={2023},
  volume={39},
  number={3},
  pages={1816-1835},
  doi={10.1109/TRO.2023.3236945}}

@INPROCEEDINGS{tao2023SEER,
  author={Tao, Yuezhan and Wu, Yuwei and Li, Beiming and Cladera, Fernando and Zhou, Alex and Thakur, Dinesh and Kumar, Vijay},
  booktitle={2023 IEEE International Conference on Robotics and Automation (ICRA)}, 
  title={SEER: Safe Efficient Exploration for Aerial Robots using Learning to Predict Information Gain}, 
  year={2023},
  volume={},
  number={},
  pages={1235-1241},
  doi={10.1109/ICRA48891.2023.10160295}}

@ARTICLE{huang2023FAEL,
  author={Huang, Junlong and Zhou, Boyu and Fan, Zhengping and Zhu, Yilin and Jie, Yingrui and Li, Longwei and Cheng, Hui},
  journal={IEEE Robotics and Automation Letters}, 
  title={FAEL: Fast Autonomous Exploration for Large-scale Environments With a Mobile Robot}, 
  year={2023},
  volume={8},
  number={3},
  pages={1667-1674},
  doi={10.1109/LRA.2023.3236573}}

@ARTICLE{placed2023Survey,
  author={Placed, Julio A. and Strader, Jared and Carrillo, Henry and Atanasov, Nikolay and Indelman, Vadim and Carlone, Luca and Castellanos, José A.},
  journal={IEEE Transactions on Robotics}, 
  title={A Survey on Active Simultaneous Localization and Mapping: State of the Art and New Frontiers}, 
  year={2023},
  volume={39},
  number={3},
  pages={1686-1705},
  doi={10.1109/TRO.2023.3248510}}

@inproceedings{wei2022Robot,
author = {Wei, Chao and Xu, Meng and Wang, Jikai and Chen, Zonghai},
title = {Robot Exploration based on Small Areas Priority Strategy},
year = {2022},
isbn = {9781450376921},
publisher = {Association for Computing Machinery},
address = {New York, NY, USA},
doi = {10.1145/3529763.3529765},
booktitle = {Proceedings of the 3rd International Conference on Service Robotics Technologies},
pages = {7–14},
numpages = {8},
location = {Chengdu, China},
series = {ICSRT '22}
}

@INPROCEEDINGS{zhu2021DSVP,
  author={Zhu, Hongbiao and Cao, Chao and Xia, Yukun and Scherer, Sebastian and Zhang, Ji and Wang, Weidong},
  booktitle={2021 IEEE/RSJ International Conference on Intelligent Robots and Systems (IROS)}, 
  title={DSVP: Dual-Stage Viewpoint Planner for Rapid Exploration by Dynamic Expansion}, 
  year={2021},
  volume={},
  number={},
  pages={7623-7630},
  doi={10.1109/IROS51168.2021.9636473}}

@inproceedings{cao2021TARE,
  title = {TARE: A Hierarchical Framework for Efficiently Exploring Complex 3D Environments},
  shorttitle = {TARE},
  booktitle = {Robotics: Science and Systems XVII},
  author = {Cao, Chao and Zhu, Hongbiao and Choset, Howie and Zhang, Ji},
  year={2021},
  date = {2021-07-12},
  publisher = {{Robotics: Science and Systems Foundation}},
  eventtitle = {Robotics: Science and Systems 2021},
  isbn = {978-0-9923747-7-8},
  langid = {english}
}

@article{zhang2020Falco,
author = {Zhang, Ji and Hu, Chen and Chadha, Rushat and Singh, Sanjiv},
year = {2020},
month = {Apr.},
pages = {},
title = {Falco: Fast likelihood‐based collision avoidance with extension to human‐guided navigation},
volume = {37},
journal = {Journal of Field Robotics},
doi = {10.1002/rob.21952}
}

@ARTICLE{duberg2020UFOMap,
  author={Duberg, Daniel and Jensfelt, Patric},
  journal={IEEE Robotics and Automation Letters}, 
  title={UFOMap: An Efficient Probabilistic 3D Mapping Framework That Embraces the Unknown}, 
  year={2020},
  volume={5},
  number={4},
  pages={6411-6418},
  doi={10.1109/LRA.2020.3013861}}

@INPROCEEDINGS{deng2020Robotic,
  author={Deng, Di and Duan, Runlin and Liu, Jiahong and Sheng, Kuangjie and Shimada, Kenji},
  booktitle={2020 IEEE/ASME International Conference on Advanced Intelligent Mechatronics (AIM)}, 
  title={Robotic Exploration of Unknown 2D Environment Using a Frontier-based Automatic-Differentiable Information Gain Measure}, 
  year={2020},
  volume={},
  number={},
  pages={1497-1503},
  doi={10.1109/AIM43001.2020.9158881}}

@article{dang2020Graphbased,
  title={Graph-based subterranean exploration path planning using aerial and legged robots},
  author={Dang, Tung and Tranzatto, Marco and Khattak, Shehryar and Mascarich, Frank and Alexis, Kostas and Hutter, Marco},
  journal={Journal of Field Robotics},
  volume = {37},
  number = {8},
  pages = {1363-1388},  
  year={2020},
  note={Wiley Online Library}
}

@INPROCEEDINGS{cieslewski2017Rapid,
  author={Cieslewski, Titus and Kaufmann, Elia and Scaramuzza, Davide},
  booktitle={2017 IEEE/RSJ International Conference on Intelligent Robots and Systems (IROS)}, 
  title={Rapid exploration with multi-rotors: A frontier selection method for high speed flight}, 
  year={2017},
  volume={},
  number={},
  pages={2135-2142},
  doi={10.1109/IROS.2017.8206030}}

@INPROCEEDINGS{bircher2016Receding,
  author={Bircher, Andreas and Kamel, Mina and Alexis, Kostas and Oleynikova, Helen and Siegwart, Roland},
  booktitle={2016 IEEE International Conference on Robotics and Automation (ICRA)}, 
  title={Receding Horizon "Next-Best-View" Planner for 3D Exploration}, 
  year={2016},
  volume={},
  number={},
  pages={1462-1468},
  doi={10.1109/ICRA.2016.7487281}}

@INPROCEEDINGS{yamauchi1997Frontierbased,
  author={Yamauchi, B.},
  booktitle={Proceedings 1997 IEEE International Symposium on Computational Intelligence in Robotics and Automation CIRA'97. 'Towards New Computational Principles for Robotics and Automation'}, 
  title={A frontier-based approach for autonomous exploration}, 
  year={1997},
  volume={},
  number={},
  pages={146-151},
  doi={10.1109/CIRA.1997.613851}}

@article{zhang1984Fast,
author = {Zhang, T. Y. and Suen, C. Y.},
title = {A fast parallel algorithm for thinning digital patterns},
year = {1984},
issue_date = {March 1984},
publisher = {Association for Computing Machinery},
address = {New York, NY, USA},
volume = {27},
number = {3},
issn = {0001-0782},
doi = {10.1145/357994.358023},
journal = {Communications Of The ACM},
month = {Mar.},
pages = {236–239},
numpages = {4}
}

@INPROCEEDINGS{choset1995Sensora,
  author={Choset, H. and Burdick, J.},
  booktitle={Proceedings of 1995 IEEE International Conference on Robotics and Automation (ICRA)}, 
  title={Sensor based planning. I. The generalized Voronoi graph}, 
  year={1995},
  volume={2},
  number={},
  pages={1649-1655 vol.2},
  doi={10.1109/ROBOT.1995.525511}}

\end{document}